\documentclass[twoside,twocolumn]{article}

\usepackage[round]{natbib}

\usepackage{amssymb,graphicx,url,amsmath,subfigure}
\usepackage{times}
\usepackage{tikz}
\usepackage{soul}
\usepackage{color}
\usepackage{xcolor}
\usepackage{algorithm}
\usepackage{algorithmic}
    \usepackage{enumerate}
\usepackage{comment}
\usepackage{amsthm}
\usepackage{diagbox}
\usepackage{colortbl}
\usepackage[linguistics]{forest}

\usetikzlibrary{shapes.geometric,positioning}

\newcommand{\Exp}{\mathbf{E}}

\newcommand{\R}{{\mathbb R}}

\newcommand{\Z}{{\mathbb Z}}

\newcommand{\cX}{{\cal X}}
\newcommand{\cY}{{\cal Y}}
\newcommand{\cW}{{\cal W}}

\newcommand{\bx}{{\bf x}}
\newcommand{\by}{{\bf y}}
\newcommand{\bX}{{\bf X}}

\newcommand{\cF}{{\cal F}}

\newtheorem{Theorem}{Theorem}

\newtheorem{Definition}{Definition}

\newtheorem{Postulate}{Postulate}

\definecolor{MyDarkGreen}{rgb}{0.17,0.46,0.25} 
\definecolor{MyDarkRed}{rgb}{0.88,0.22,0.21} 
\definecolor{MyDarkBlue}{rgb}{0.11,0.11,0.70}

\definecolor{lightgray}{gray}{0.85}
\sethlcolor{Dandelion}

\usetikzlibrary{arrows,shapes,plotmarks,positioning}
\usetikzlibrary{decorations.markings}

\newdimen\arrowsize
\pgfarrowsdeclare{arcsq}{arcsq}
{
  \arrowsize=0.2pt
  \advance\arrowsize by .5\pgflinewidth
  \pgfarrowsleftextend{-4\arrowsize-.5\pgflinewidth}
  \pgfarrowsrightextend{.5\pgflinewidth}
}
{
  \arrowsize=1.5pt
  \advance\arrowsize by .5\pgflinewidth
  \pgfsetdash{}{0pt} 
  \pgfsetroundjoin   
  \pgfsetroundcap    
  \pgfpathmoveto{\pgfpoint{0\arrowsize}{0\arrowsize}}
  \pgfpatharc{-90}{-140}{4\arrowsize}
  \pgfusepathqstroke
  \pgfpathmoveto{\pgfpointorigin}
  \pgfpatharc{90}{140}{4\arrowsize}
  \pgfusepathqstroke
}

\tikzset{>=stealth'} 
\tikzstyle{graphnode} = 
   [circle,draw=black,minimum size=22pt,text centered,text
     width=22pt,inner sep=0pt] 
\tikzstyle{var}   =[graphnode,fill=white]
\tikzstyle{vardashed}   =[graphnode,draw=gray,fill=white]
\tikzstyle{obs}   =[graphnode,fill=black,text=white]
\tikzstyle{obsgrey}   =[graphnode,draw=white,fill=lightgray,text=black]
\tikzstyle{par}    =[graphnode,draw=white,fill=red,text=black] 
 \tikzstyle{crucial} =[graphnode,draw=white,fill=yellow,text=black] 
\tikzstyle{fac}   =[rectangle,draw=black,fill=black!25,minimum size=5pt]
\tikzstyle{facprior} =[rectangle,draw=black,fill=black,text=white,minimum size=5pt]
\tikzstyle{edge}  =[draw=white,double=black,very thick,-]
\tikzstyle{blueedge}  =[draw=white,double=blue,very thick,-]
\tikzstyle{rededge}  =[draw=white,double=red,very thick,-]
\tikzstyle{prior} =[rectangle, draw=black, fill=black, minimum size=
5pt, inner sep=0pt]
\tikzstyle{dirprior} = [circle, draw=black, fill=black, minimum
size=5pt, inner sep=0pt]

\tikzstyle{dot_node}=[draw=black,fill=black,shape=circle]

\tikzset{every picture/.style={line width=0.6pt}}
\tikzset{every picture/.style={outer sep=.4mm}}
\tikzstyle{graphnode} = 
   [circle,draw=black,minimum size=25pt,text centered,inner sep=0.2mm] 
\tikzstyle{observed}   =[graphnode,fill=white,text=black]
\tikzstyle{unobserved}   =[graphnode,fill=white,text=black,style=dashed]
\tikzstyle{graphnodesmall} = 
   [circle,draw=black,minimum size=20pt,text centered,inner sep=0.2mm] 
\tikzstyle{observedsmall}   =[graphnodesmall,fill=white,text=black]
\tikzstyle{unobservedsmall}   =[graphnodesmall,fill=white,text=black,style=dashed]
\tikzstyle{graphnodestiny} = 
   [circle,draw=black,minimum size=15pt,text centered,inner sep=0.2mm] 
\tikzstyle{observedtiny}   =[graphnodestiny,fill=white,text=black]
\tikzstyle{unobservedtiny}   =[graphnodestiny,fill=white,text=black,style=dashed]

\date{November 24, 2021}

\begin{document}

\title{Causal versions of  Maximum Entropy \\ and Principle of Insufficient Reason} 

\author{Dominik Janzing\\
{\small Amazon Research T\"ubingen, Germany }\\
{\small janzind@amazon.com} }

\maketitle

\begin{abstract}
The Principle of Insufficient Reason (PIR) assigns equal probabilities to each alternative of a random experiment whenever 
there is no reason to prefer one over the other. The Maximum Entropy Principle (MaxEnt) generalizes PIR to the case where statistical information like expectations are given. It is known that both principles result in paradoxical probability updates for  
joint distributions of cause and effect. This is because constraints on the conditional
 $P({\rm effect | cause})$ result in changes of $P({\rm cause})$ that assign higher probability to those values
 of the cause that offer more options for the effect, suggesting
'intentional behaviour'. 
 Earlier work therefore suggested
sequentially maximizing (conditional) entropy according to the causal order, but without further justification apart from plausibility on toy examples. We justify causal modifications of PIR and MaxEnt 
by separating constraints into restrictions for the cause and restrictions for the mechanism that generates the effect from the cause.  
We further sketch why Causal PIR also entails 'Information Geometric Causal Inference'. 

We briefly discuss problems of generalizing the causal version of MaxEnt to arbitrary causal DAGs.    
\end{abstract}

\section{Introduction}

Understanding asymmetries between cause and effect has attracted researchers 
from the field of causal discovery particularly  since two decades. 
One challenging problem motivated by the goal of understanding these asymmetries is to 
 distinguish cause and effect from their bivariate distribution. This task cannot be solved by causal discovery methods that rely on conditional independences only \citep{Spirtes1993,Pearl:00}, but new approaches 
 employ statistical properties other than conditional independences. 
 They rely, for instance, on the additive noise assumption \citep{Kano2003,Hoyer,Mooij2016} or a generalization of the latter \citep{Zhang_UAI}, or
   on asymmetries with respect to some notion of description complexity \citep{Algorithmic,Marx2017,Kocaoglu2017},  or differences regarding regression error 
   \citep{Bloebaum17}. For an overview see also \citet{causality_book} and \citet{Guyon2019}, but also \citet{misconceptions} for a critical discussion of  some ideas.  Although distinction of cause and effect from purely observational data is still challenging,
 these approaches have stimulated discussions in various directions regarding inferential asymmetries of cause and effect.
 On the one hand, the relation to the arrow of time in physics has been described by \citet{MitArmen} and \citet{AICarrowoftime}.
 On the other hand, it has been argued that the asymmetries entail implications for machine learning
for scenarios where the causal direction is known \citep{anticausal,Bengio2019}.  

Here we describe an asymmetry between cause and effect with respect to how we assign priors to a set of possible outcomes of an experiment. 
Among the most prominent principles to assign priors is the 'Principle of Insufficient Reason' (PIR) and the Principle of Maximum  Entropy (MaxEnt) \citep{Jaynes2003}.
PIR assigns uniform probabilities to a set of possible outcomes whenever the knowledge about the outcomes is
invariant under permutations. MaxEnt, which generalizes PIR, chooses a prior that maximizes entropy subject to the known constraints. 
For case where the causal direction is known, \citet{SunLauderdale}  have argued that MaxEnt can result in implausible distributions and more natural joint distributions
result from a sequential maximization: first maximize entropy of the cause subject to all constraints relevant for the latter, and then  
the conditional entropy of the effect, given the cause, subject to all remaining constraints. However,  the arguments of \citet{SunLauderdale} were merely based on intuition without further justification.

On a related note, \citet{Ziebart2013} propose a 'maximum causal entropy principle' for a scenario with two interacting processes $X_t,Y_t$ where $X_t$ is known and $Y_t$ is inferred from its own past and from $X_t$ and its past
via a sequential maximization of conditional entropy. 
\citet{Ziebart2013} justify the sequential update by arguing that constraints that involve future observations 
should be ignored {\it at that respective point in time}. In the appendix we argue that this justification is not sufficient for our purpose.

The goal of this paper 
is to derive the sequential maximum entropy update rule proposed by  \citet{SunLauderdale}  from principles that we consider slightly more basic. 
To this end, Section \ref{sec:PIR} 
discusses a simple scenario suggesting that also PIR requires the same modification as MaxEnt.
Section \ref{sec:mech} tries to justify {\it Causal PIR} from a deeper principle of independent mechanisms, but also raises questions that remain open in this regard. Section \ref{sec:maxent} derives the causal version of MaxEnt by \citet{SunLauderdale}
from applying Causal PIR to {\it empirical} distributions. 
Section \ref{sec:N} describes some problems of generalizing Causal MaxEnt to arbitrary causal DAGs.
Section \ref{sec:igci} shows that Information Geometric Causal Inference \citep{deterministic} can be derived from Causal PIR similar to Causal MaxEnt. 

Proposing new practical inference rules is beyond the scope of this paper. Instead, it aims at better understanding relations between asymmetries of cause versus effect described earlier.


\section{Causal Version of PIR  \label{sec:PIR}} 

\subsection{Standard PIR}  

The 'Principle of Insufficient Reason (PIR)', also called 
'Laplace's Principle of Insufficient Reason' or 'Principle of Indifference' \citep{Jaynes2003},
states that in the absence of any relevant evidence, agents should distribute their credence (or 'degrees of belief') equally among all the possible outcomes under consideration. More explicitly, PIR advices to 
consider all possible alternatives in a random experiment equally likely. 
For the simple example where we know that one of $n$ urns contains a ball, PIR considers each of the urns as an  equally likely location and assigns $P(j)=1/n$ to each case $j=1,\dots,n$. 
For a discussion of justifications of PIR we refer to \cite{Uffink1995}, where also the relation to MaxEnt is discussed in detail.

For our purpose, it is also instructive to 
rephrase PIR by stating that it advices the uniform prior 
whenever there is no evidence that breaks the {\it symmetry} between the alternative outcomes. 
In a way, PIR then gets a circular structure because any argument against the uniform prior implicitly raises doubts about the symmetry of the problem (obviously, the uniform distribution is the only one that
is symmetric under permutation of the alternatives). One insight of our discussion below will be that a reasonable use of PIR is not symmetric with respect to interchanging cause and effect. 
We are agnostic about whether one should consider this merely as an advice on how to {\it properly apply PIR in a cause-effect scenario} or as a {\it causal modification of PIR}.

\subsection{Motivating Causal PIR for a simple mechanical device}  
\begin{figure}
\centerline{
\resizebox{5cm}{!}{%
\begin{tikzpicture}
\draw (-.25,0) -- (-.25,2);     
\draw (.25,0) -- (.25,2);       
\draw (1.75,0) -- (1.75,2);   
\draw (2.25,0) -- (2.25,2);   
\draw (3.75,0) -- (3.75,2);   
\draw (4.25,0) -- (4.25,2);   
\draw (.75,4) -- (.75,6);     
\draw (1.25,4) -- (1.25,6);    
\draw (2.75,4) -- (2.75,6);   
\draw (3.25,4) -- (3.25,6);   
\draw (4.75,4) -- (4.75,6);   
\draw (5.25,4) -- (5.25,6);   
\draw (-.25,2) -- (.75,4);
\draw (2.25,2) --  (1.25,4);
\draw (3.75,2) -- (2.75,4);
\draw (4.25,2) -- (5.25,4);
\draw (0.25,2) -- (1,3.5) ; 
\draw (4,2.5) -- (4.75,4); 
\draw (4,2.5) -- (3.25,4); 
\draw (1.75,2) -- (1,3.5); 
\node [text width=0.5cm] at (-1,1) {$Y=$};
\node [text width=0.5cm] at (0.2,1) {$3$};
\node [text width=0.5cm] at (2.2,1) {$2$};
\node [text width=0.5cm] at (4.2,1) {$1$};
\node [text width=0.5cm] at (-1,5) {$X=$};
\node [text width=0.5cm] at (1.2,5) {$1$};
\node [text width=0.5cm] at (3.2,5) {$2$};
\node [text width=0.5cm] at (5.2,5) {$3$};
\shade[ball color = gray!40, opacity = 0.4] (3,8) circle (0.2cm);
\draw [->] (3,7.5) -- (3,7);
\end{tikzpicture}
}
}
\caption{\label{fig:ball} A ball enters our mechanical device from the top. Without additional information, we would consider all three options ($X=1,X=2,X=3$) equally likely, that is, assign the probability $1/3$ to them. This results in probability $2/3$ for $Y=1$ and probability $1/6$ each for $X=2$ and $X=3$.} 
\end{figure}
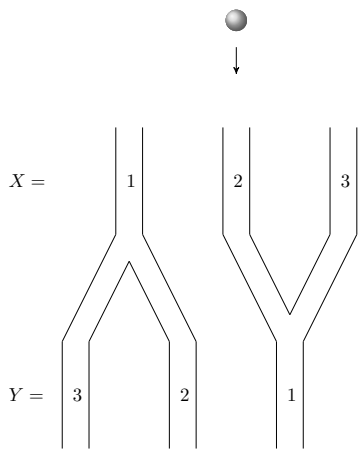

Consider the mechanical device depicted in Figure~\ref{fig:ball}. It consists of a system with channels having three different entries (top of the figure) and three exits (bottom).  The first entry splits into two different channels, while
the second and the third entry lead to the same exit. Let us label the three entries with the variable $X$ attaining the values $1,2,3$, while $Y$ labels the exits $1,2,3$. 
Assume we know that a ball enters one of the three entrances at the top. In absence of any further information, we would consider all three options as equally likely, that is $P(X=1)=P(X=2)=P(X=3)=1/3$, in agreement with PIR. 
When rolling through the channel, the ball will take one of the three exits. Whenever it entered at the entrance $2$ or $3$, it can only take the exit $Y=1$ due to the topology of the channels. In case it entered at entrance $1$, it has the two options 
later, namely exits $Y=2$ or $Y=3$. We now apply PIR for the {\it conditional} distribution of $Y$ given $X$ and assume that both alternatives are equally likely. 
The scenario thus yields  the joint probabilities
shown in the table in Figure~\ref{fig:xy}, left. This distribution is clearly asymmetric with respect to $X$ and $Y$ although the topology of the channels is symmetric. Assuming that the ball enters from the bottom, that is, $Y$ labels the entrances and $X$ the exits, thus induces the joint distribution in the table in Figure~\ref{fig:xy}, right, which is obtained by swapping the roles of $X$ and $Y$. 

Lead by our intuition, we have applied PIR twice: first for $X$, and then for $Y$, given $X$. However, the simplicity of the scenario blurs a non-trivial step in this way of reasoning, namely that the experiment is not symmetric with respect to time inversion, or, 
which is equivalent here, with respect to swapping cause and effect. Here, our asymmetry of reasoning is implicitly based
on a belief about the difference between cause vs. effect and past vs. future.

Note that our mechanical toy example does not describe the typical scenario of cause-effect inference
since it is uncommon to know the mechanism that relates cause and effect, that is, only the direction is unknown. 
Typically, we are given observations from $X,Y$ instead of knowledge on the mechanisms.
 Yet the example is helpful to motivate Causal PIR, which is later used to motivate Causal MaxEnt, which, in turn, is relevant for more realistic inference scenarios.

\subsection{Fallback to standard PIR when causal direction is unknown}  
To elaborate on this, note that the topology of the channel allows
$4$ different $x,y$-pairs. 
Without knowing whether the ball enters from the top or the bottom, PIR lets us assign equal probabilities to each of them
since the device is symmetric once the knowledge of the direction of the motion is lost.  
Obviously, the symmetry of the problem now results in the distribution shown in the table in Figure~\ref{fig:xy}, middle. Note that this distribution may not only be natural when we are {\it agnostic} about the
causal direction, but also if neither of the causal directions is  true and the relation between $X$ and $Y$ is due to a common cause. Although the following scenario may seem less natural than the first two ones with $X$ or $Y$ as cause, we mention it to cover also the common cause scenario. 
Assume that the ball drops from the sky into one of the channels and lies there at some point at rest. If it lies in the regions 
$X=2,3$ or $Y=2,3$, its position already defines a unique $(x,y)$-pair since these values can only occur together with
a unique value of $Y$ or $X$, respectively. In the case where it lies in the regions
$X=1$ or $Y=1$, we push it towards the branching point to generate the corresponding random value of $Y$ or $X$, respectively.  This way, we have again generated a scenario in which we have no reason to prefer any of the $4$ possible $(x,y)$-pairs over the other.
One can argue that the causal structure of this scenario is the DAG shown in Figure~\ref{fig:xy}, middle, where
some 'big' unobserved variable $Z$ affects both $X$ and $Y$, where $Z$  contains position and momentum of the ball and the noise which determines the branching process.

\subsection{Paradoxes with standard PIR \label{subsec:paradox}} 
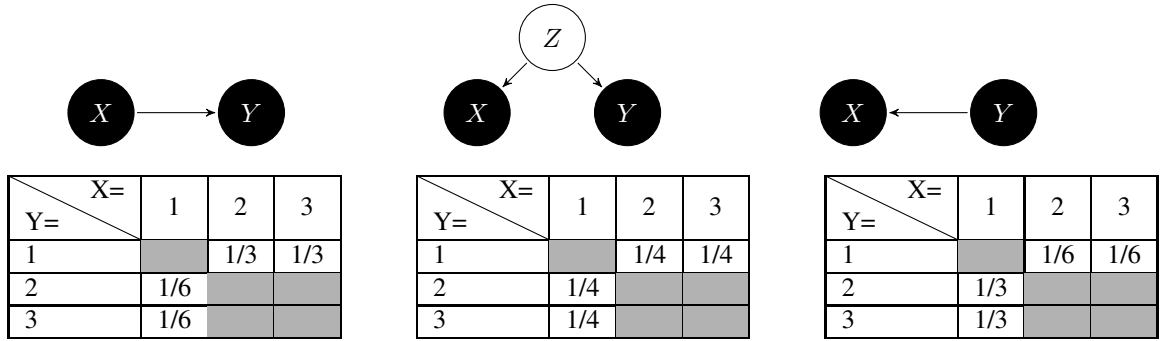
\begin{figure*}
\hspace{1.3cm}
 \begin{tikzpicture}
    \node[obs] at (0,1) (X) {$X$};
    \node[obs] at (2,1) (Y) {$Y$}  edge[<-] (X);    
    \node[obs] at (5,1) (Xc) {$X$} ;
    \node[var] at (6,2) (Z) {$Z$} edge[->] (Xc) ;  
    \node[obs] at (7,1) (Yc) {$Y$} edge[<-] (Z) ;
    \node[obs] at (10,1) (Xa) {$X$};
    \node[obs] at (12,1) (Ya) {$Y$}  edge[->] (Xa);
  \end{tikzpicture}
\[
\begin{tabular}{|l|c|c|c|}
\hline
\diagbox{Y=}{X=} & 1 & 2 & 3\\
\hline
  1  &  \cellcolor[gray]{0.7}  &    1/3  &   1/3  \\
\hline
  2 &  1/6  & \cellcolor[gray]{0.7}   &  \cellcolor[gray]{0.7}    \\
\hline
  3 & 1/6   &  \cellcolor[gray]{0.7}   &  \cellcolor[gray]{0.7}  \\
  \hline
\end{tabular}  
\hspace{1cm}
\begin{tabular}{|l|c|c|c|}
\hline
\diagbox{Y=}{X=} & 1 & 2 & 3\\
\hline
  1  & \cellcolor[gray]{0.7}   &   1/4 &  1/4\\
\hline
  2 &  1/4   &  \cellcolor[gray]{0.7}  &  \cellcolor[gray]{0.7}  \\
\hline
  3 & 1/4   &  \cellcolor[gray]{0.7}   &  \cellcolor[gray]{0.7} \\
  \hline
\end{tabular}  
\hspace{1cm}
\begin{tabular}{|l|c|c|c|}
\hline
\diagbox{Y=}{X=} & 1 & 2 & 3\\
\hline
  1  &  \cellcolor[gray]{0.7}  &   1/6  &  1/6\\
\hline
  2 &  1/3  & \cellcolor[gray]{0.7}   &   \cellcolor[gray]{0.7}  \\
\hline
  3 & 1/3  &   \cellcolor[gray]{0.7}  &  \cellcolor[gray]{0.7}  \\
  \hline
\end{tabular}  
\]
\caption{\label{fig:xy} Left: Joint probabilities when the ball enters from the top ($X$ is the cause). Middle: probabilities when the ball enters from the sky (common cause). Right: when the ball enters from the bottom ($Y$ is the cause).}
\end{figure*}

As the table in Figure \ref{fig:xy}, middle, shows,
assigning equal probabilities to all $4$ possible cases result in higher probabilities to those values of the cause
that admit more options for the effect -- which suggests 'intentional behaviour'. Note, however, that the latter interpretation
is prone of confusing {\it ontic} and {\it  epistemic} perspectives:  whenever the restriction to these $4$ alternatives
comes from our knowledge {\it about the underlying mechanism connecting the cause $X$ with the effect $Y$}, it is indeed
irrational to consider $x$-values more likely for which there exists a larger number of possible $y$-values later. However, if we know, for some other reason, that $(x,y)$ is one of the above $4$ cases (e.g. because someone told us without telling us $(x,y)$),
there is nothing wrong with updating our {\it subjective prior}  for $X$ in the way resulting from the uniform distribution over the $4$ possible pairs. After all, this Bayesian update entails no statement on the underlying causal mechanism. 
This distinction will be further discussed in Section \ref{sec:mech}, where we also mention open problems regarding ontic versus epistemic interpretation of constraints.  

Similar paradoxes with standard PIR have been described by \citet{Hunter1986,Hunter1989} in a critical
discussion of MaxEnt. 
He described a scenario which he called 'Pearl's Puzzle'\footnote{Hunter writes: ``The example was given by personal communication and has been floating around the uncertain reasoning community for sometime. Pearl informs me that the example was discovered by Norman Dalkey but was first taken as a counterexample to MaxEnt by Pearl''}, which we briefly sketch. 
Assume three individuals $A,B,C$ are invited to a party but don't know who will be joining. 
Further assume we consider, a priori, all $8$ possible combinations equally likely. In addition, we know that $A,B$ decide independently of each other and of $C$ whether they join, but $C$ will call the host to ask whether both $A$ and $B$   
have accepted the invitation and stay at home in this case to avoid seeing both of them together.
After accounting for this extra information (excluding the case where $A,B,C$ occur), we are left with $7$
remaining combinations, which we would assign equal probabilities to.
According to such an update, the joint distribution of $A,B$ has changed after accounting for the information that $C$'s decision depends on $A$ and $B$. Phrasing it in causal terms, the puzzle reads as follows: 
$A,B$ are the causes and  $C$'s behaviour their effect. Learning about how $C$'s decision depends on $A$ and $B$
actually  changes the belief about the mechanism according to which the effect depends on its causes. It is disturbing that
an update on this mechanism affects the distribution of the causes (one can also show, which Pearl describes as the main puzzle, that $A$ and $B$ even become dependent by this update). 

We will later elaborate on this in the context of the so-called Principle of Independent
Mechanisms \citep{causality_book}, since Hunter's and Pearl's discussions are  already lead by such an independence assumption. 

To conclude with 'Pearl's puzzle' we briefly sketch how it gets resolved by a sequential use of PIR: since $A$ and $B$ are the causes, we assign a uniform
prior over all $4$ possible truth values. Afterwards, we assign a uniform prior over all remaining options for $C$: whenever $A$ and $B$ are coming, 
$C$ stays at home with probability $1$, for all other cases he would decide to come with probability $1/2$. By construction, whether or not $C$ is coming,
is irrelevant for $A$ and $B$. 

\subsection{General definition of Causal PIR} 

\begin{figure}
\centerline{
\resizebox{5cm}{!}{%
\begin{tikzpicture}
\draw (-.25,0) -- (-.25,2);     
\draw (.25,0) -- (.25,2);       
\draw (1.75,0) -- (1.75,2);   
\draw (2.25,0) -- (2.25,2);   
\draw (3.75,0) -- (3.75,2);   
\draw (4.25,0) -- (4.25,2);   
\draw (.75,4) -- (.75,6);     
\draw (1.25,4) -- (1.25,6);    
\draw (2.75,4) -- (2.75,6);   
\draw (3.25,4) -- (3.25,6);   
\draw [line width=0.5cm,gray] (3,3.9) -- (3,6);
\draw (4.75,4) -- (4.75,6);   
\draw (5.25,4) -- (5.25,6);   
\draw (-.25,2) -- (.75,4);
\draw (2.25,2) --  (1.25,4);
\draw (3.75,2) -- (2.75,4);
\draw (4.25,2) -- (5.25,4);
\draw [line width=0.5cm,gray] (4,0) -- (4,2.1);
\draw [line width=0.5cm,gray] (4,2) -- (3,4);
\draw (0.25,2) -- (1,3.5) ; 
\draw (4,2.5) -- (4.75,4); 
\draw (4,2.5) -- (3.25,4); 
\draw (1.75,2) -- (1,3.5); 
\node [text width=0.5cm] at (-1,1) {$Y=$};
\node [text width=0.5cm] at (0.2,1) {$3$};
\node [text width=0.5cm] at (2.2,1) {$2$};
\node [text width=0.5cm] at (4.2,1) {$1$};
\node [text width=0.5cm] at (-1,5) {$X=$};
\node [text width=0.5cm] at (1.2,5) {$1$};
\node [text width=0.5cm] at (3.2,5) {$2$};
\node [text width=0.5cm] at (5.2,5) {$3$};
\end{tikzpicture}
}
}
\caption{\label{fig:path}  For the gray path with unknown direction, it is more likely that the ball entered from the top than from the bottom according to Causal PIR.} 
\end{figure}
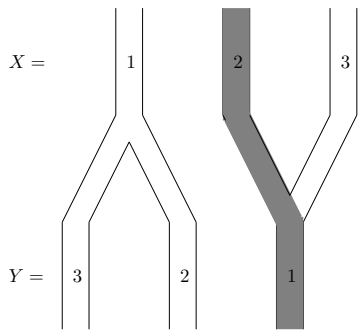

The way we defined the joint distribution for the mechanical device can be described by the following principle, which
also solved the above `puzzle':  
\begin{Definition}[Causal PIR]\label{def:causalPIR} 
Let $X$ and $Y$ be cause and effect with values in finite sets $\cX$ and $\cY$, respectively. If the only knowledge about
an observation $(x,y)$ is that it lies in some subset 
 $S\subset \cX \times \cY$,  
Causal PIR assigns uniform distribution to all
possible $x$, for which there exists an $y$ such that $(x,y)\in S$.
Then causal PIR assigns the uniform prior over all remaining options for $y$, given $x$ (that is, all $y$ for which
$(x,y)\in S$).
\end{Definition}
A priori, we have introduced Causal PIR only as a principle for constructing a prior when the causal direction is known. Conversely, one can certainly use its asymmetry to infer the causal direction by preferring the one with larger likelihood:
\begin{Definition}[Causal PIR based cause-effect inference]\label{def:MLCausalPIR} Given an observation $(x,y)$ generated by either the causal structure $X\to Y$ or $Y\to X$. Infer that the true causal direction is the one for which $(x,y)$ has larger likelihood
according to the Causal PIR prior.
\end{Definition} 
 Observing, for instance, the path in Figure \ref{fig:path}, we thus infer that the ball entered from the top rather than from the bottom: we obtain likelihood $1/3$ for the former versus $1/6$ for the latter. 
In a more informal way, we state Causal PIR as follows:
\begin{Postulate}[informal version of Causal PIR]
Prefer causal models (i.e. directions, DAGs, structural equations) for which
\\
$\bullet$ the observed effect admits many values of the cause and
\\
$\bullet$ the observed cause admits few values of the effect.
\end{Postulate}

\section{Independent mechanism update \label{sec:mech}}

This section and Section \ref{sec:maxent} repeatedly refer to the Principle of Independent Mechanisms (IM), which we
briefly introduce for the special case of a cause-effect pair.
A priori, IM is an informal principle stating  that, for an unconfounded cause-effect relation, there  
should be two independent mechanisms in place, one that generates the cause and one that generates the effect from the cause
(see \cite{causality_book}, Section 2.1, for an overview and discussion of its different aspects). 
IM has been used as foundational justification for cause-effect inference. One formalization of IM in the literature is the Algorithmic Independence of
Conditionals (AIC) by \citet{Algorithmic} and \citet{LemeireJ2012}, stating that the shortest description of $P_{X,Y}$ is given by separate descriptions of $P_X$ and $P_{Y|X}$.\footnote{Further concrete conceptualizations of IM are:
(1) the hypothesis that unlabelled data in semi-supervised learning is only helpful for a so-called `anticausal prediction' scenario (where the cause is predicted from the effect), while it is pointless for `causal prediction' (when the effect is predicted from the cause), see \citep{anticausal}. Further, IM has been formalized as  (2) uncorrelatedness of the logarithmic slope of the function connecting $X$ and $Y$ 
with the density of $X$ in `Information Geometric Causal Inference' \citep{deterministic}, and (3) uncorrelatedness of the (absolute squared) transfer function connecting $X$ and $Y$ with the power spectrum of $X$ 
when $X$ and $Y$ are time series connected by a linear filter (`Spectral Independence Criterion' by \citet{sic_icml}).}
This version will  be relevant in Section \ref{sec:maxent}, while the following subsection will interpret IM  
in the sense of decomposing the constraint $S$ according to one that refers to the cause and one referring to the relation between cause and effect.

\subsection{Constraints on the cause and constraints on functions \label{subsec:priorf}}
We will now describe a principle that justifies Causal PIR in Definition \ref{def:causalPIR}.  
To this end, let $\cF:=\cY^\cX$ denote the set of functions  $f:\cX \to \cY$ and  define the formal random variable $F$ attaining values $f\in \cF$. Note that $\cF$ can be represented as 
the $k$-fold cartesian product of $\cY$  with $k:=|\cX|$, whose components are indexed by $x\in \cX$. In other words, a function $f$ is represented
by the $k$-tuple $(f(x^1),\dots,f(x^k))$ if $\cX=\{x^1,\dots,x^k\}$.  
Accordingly, a distribution on $\cF$ is a joint distribution on this cartesian product. Its marginal distribution on component $x$ describes the conditional
probability $P_{Y|X=x}$ (see, for instance, \cite{causality_book}, Section 3.4). 

Let $P_\cF$ be the uniform distribution\footnote{Note that Hunter's solution of `Pearl's puzzle' \citep{Hunter1989} also uses an update of a distribution over functions ('probability measures over counterfactuals'), but is based on the assumption that the constraint is known to refer to the function only, while our scenario describes a constraint for $(x,y)$ for which it is not a priori known what it tells us about the function.} on $\cF$. Fortunately, the uniform distribution 
has product structure over the components of the cartesian product, which renders $P_\cF$ 
particularly easy to deal with. Further, for every input $x$, every output $y$ is equally likely. In other words, 
the uniform prior over all functions induces conditional distributions $P_{Y|X=x}$  that are uniform for all $x$.
After assuming also a uniform prior $P_X$, we have thus obtained a uniform prior $P_{X,Y} = P_X P_{Y|X}$ 
over all $|\cX | |\cY|$ combinations, which are $9= 3\cdot 3$ in the above example.  

After having defined our prior on $\cX$ and $\cF$, let us obtain the additional information
that the entire device only generates $(x,y)$-pairs in some set $S \subset \cX \times \cY$. 
For the example above, these are the  $4$ combinations shown in the tables in Figure~\ref{fig:xy}. 
We now assume that the constraint $S$ is the result from two independent mechanisms, one for $\cX$ and one for $\cF$:
\begin{Postulate}[separation of constraints]\label{post:sep} 
Given the constraint  $(x,y)\in S$ for a cause-effect pair $(X,Y)$, we 
assume, by default, that this constraint in enforced by two separate
mechanisms. First, there is a mechanism that enforces all $x$ to be in the set 
\[
S_X := \{ x\in \cX \, | (x,y) \in S \hbox{ for some } y\in \cY \}. 
\] 
Second, there is a mechanism that enforces functions to be in the set 
\[
S_F := \{ f \in \cF  \, | (x,f(x))\in S  \hbox{ for all } x\in S_X \}. 
\]
\end{Postulate}
To better understand the postulate it helps to say what kind of mechanisms it {\it excludes}: 
Imagine an agent who chooses functions $f$ that violate $(x,f(x))\in S$ for {\it some} inputs $x\in S_X$,
but always makes sure that these functions are only combined with inputs $x$ for which the constraints are satisfied. 
In other words, the agent ensures $(x,f(x))\in S$ by combining $x$ and $f$ in a smart way.  In this case, we would say that
the mechanism choosing $x$ and the mechanism choosing $f$  are {\it dependent}. In Subsection \ref{subsec:just} we will discuss in what sense  this would violate the Principle of
Independent Mechanisms.

The above restrictions for $X$ and $F$ together generate the restriction $S$.
Note that the above separation of $S$ into $(S_X,S_F)$ entails {\it minimal commitment} on both components $x$ and $f$  (while still preserving independence)  in the following sense.
First, it is obvious that no proper superset $\tilde{S}_X \supset S_X$ guarantees that $(x,y)\in S$, regardless of the constraints for the functions.  Second, no larger set $\tilde{S}_F \supset S_F$ guarantees  $(x,y)\in S$ unless
we require $x$-values and functions $f$ to respect {\it joint} constraints.

For our mechanical device above, the constraint for $X$ reads that there are $3$ possible entries $X=1,2,3$. 
While this constraint is trivial since our set $\cX$ contains only these $3$ values, one could also think of a set $\cX$ that is a priori larger
until our information on $S$ restricts the options for $X$ to the subset $S_X$ consisting of these $3$ values.  The constraints on $F$ that we conclude from the joint constraint $S$
consists in excluding all functions that map $X=1$ to $y$-values other than $Y=2,3$ and $X=2,3$ to values other than $1$.
Nevertheless,  Postulate \ref{post:sep}  is less innocent than our toy example suggests. 
We will therefore further discuss its justification in Subsection \ref{subsec:just}.

We now obtain the following technically simple 	result, which we phrase as a theorem since
it considers Causal PIR as an implication of the more basic Postulate \ref{post:sep}:   
\begin{Theorem}[Causal PIR from independent mechanism update]\label{thm:main}	       
Let $P_X$ and $P_\cF$ be uniform distributions on $\cX$ and $\cF$, respectively. Then 
the conditional distribution of  $P^S_X$ is the uniform distribution over all $S_X$. 
Further, for every $x\in S_X$, the conditional  $P^S_{Y|X=x}$ resulting from the conditional distribution $P^S_F$ 
 is the uniform distribution over all
 $y$ for which $(x,y)\in S$.
\end{Theorem}
\proof{The first statement is obvious. 
To show that the conditional is also uniform over all remaining options, we represent each function
$f$ as the $k$-tuple of $y$-values 
\[
(y^1,\dots,y^k):=(f(x^1),f(x^2),\dots,f(x^k)),
\]
 where $x^1,\dots,x^k$ denote the elements of $S_X$.
The uniform prior over all function thus amounts to the uniform prior over all $k$-tuples
$(y^1,\dots,y^k) \in \cY^k$. Since the uniform distribution over a cartesian product factorizes over its components,
we can perform the update independently for each $x^j$ and obtain a uniform distribution
over all $y$ for which $(x^j,y) \in S$.  $\square$  
}

\subsection{Justifying separation of constraints \label{subsec:just}}
The remaining subsection is devoted to the justification of Postulate \ref{post:sep}. It needs to be informal because it is a discussion on beliefs about the world rather than insights from statistics or any other branches of mathematics. Further, it can be seen as  `abstract physics' in which the `hardware' of the underlying processes is unspecified. We also briefly mention relations to the thermodynamic arrow of time and thus reach a domain that goes beyond the scope of this paper. Accordingly, further justification of Postulate \ref{post:sep} could also raise questions of theoretical physics. However, the main focus on this Subsection is the question {\it which implicit further assumptions 
are made} when 
Postulate \ref{post:sep} is said to be entailed by the Principle of Independence Mechanisms.

\paragraph{Constraints from knowledge versus constraints from mechanisms}
The first distinction we need to make regarding the constraint $(x,y) \in S$ is whether we assume that there is a 
{\it mechanism} that generates pairs in $S$ or whether we know that a particular experiment resulted in an $(x,y)$-pair in $S$ {\it by chance}
(recall our remarks regarding {\it ontic} versus {\it epistemic} perspectives in Subsection \ref{subsec:paradox}). 
In the second case, Postulate \ref{post:sep} does not make sense: if $(x,y)\in S$ is not the result of mechanisms that enforces
$(x,y)$ to lie in $S$, it is pointless to postulate {\it separate} mechanisms.  In this case, the further argument resulting in 
Theorem \ref{thm:main} breaks down: 
 There is, a priori, no reason why {\it our knowledge} about
$X$ and $F$ could not render them dependent,\footnote{In general, knowledge about a pair of events can result in a subjective prior that renders them dependent although they are not causally related, as emphasized also by \citet{Jaynes2003}.}  although the Principle of Independence Mechanisms
 states that the true mechanisms contain no information about each other. 
 --Note that \citet{Algorithmic} formalize {\it independence}  via {\it algorithmic} information, 
 which is an ontic perspective since it relies on the description length of {\it known} mechanisms. 

One can certainly also justify an {\it epistemic} 'Principle of Independence Mechanisms' stating that our prior
about cause-effect pairs should factorize between the mechanism generating the cause and the mechanism generating the effect from the cause, but the factorization breaks down  after  joint observations from cause and effect are available.
Although this raises doubts about  Postulate \ref{post:sep}, we now discuss what kind of inductive bias provides further support. 

\paragraph{Bias for ontic interpretation}
Let us now describe a scenario in which knowing $(x,y)\in S$ does provide evidence {\it for the presence of a mechanism} that enforces or at least supports outcomes in $S$. To this end, assume that the sets $\cX$ and $\cY$ are huge (e.g. binary words of length $n$ with $n\geq 100$). Further,  assume that $S$ is a {\it strong}  constraint in the sense that it allows only a small fraction of possible outcomes, that is, $|S|\leq (|\cX| \times |\cY|)/k$ for some 
huge number $k$.
Assuming, a priori, a uniform distribution $P_{X,Y}$ on  $|\cX| \times |\cY|$, that is, we have 
$
P\{ (X,Y) \in S\} \leq \frac{1}{k}. 
$ 
Given some fixed $S$ with this property, we would certainly argue that an observation in $S$ 
is unlikely without a mechanism that increases the probability of outcomes in $S$. 
At first glance, it seems that such a conclusion is only possible if
$S$ has been defined {\it prior} to the experiment. However, it still holds when we can identify a set $S$ {\it after} observing $(x,y)$ provided that 
 $S$ has {\it low description length} (here we  formalize description length via Kolmogorov complexity \citep{Vitanyi08}, that is, let $K(S)$ denote the length of the shortest
self-delimiting program that decides whether any pair $(x,y)$ is in $S$). 
With these assumptions, it is unlikely to obtain an outcome  $(x,y)$ in {\it any} such tiny set $S$ with low complexity.
To see this, let $U$ be the union of all sets $S$ with 
$K(S)\leq \ell$ and $|S|\leq k$.  
Since the number of programs of length at most
$\ell$ is at most $2^\ell$  \citep{Vitanyi08}, the probability for obtaining a result in any of these low complexity sets can be
bounded from above as follows:
\begin{equation} \label{eq:compl} 
P\{ (X,Y) \in U \} \leq \frac{2^\ell}{k}.
\end{equation} 
In  case the right hand side of \eqref{eq:compl} is still significantly smaller than $1$ we assume that observing $(x,y)\in S$ indicates the presence of a mechanism that increases the probability 
of $S$ (compared to the uniform distribution we started with). We phrase this insight as an informal postulate:
\begin{Postulate}[bias towards mechanisms vs state of knowledge]\label{post:mech} Given a
system with finite `state space' $\cW$,
then the information that the actual state $w$ lies in some set $S$ with low complexity
(in the sense that $|S| \cdot 2^{K(S)}\ll |\cW|$) is considered as strong evidence for the presence of 
a mechanism that increases the likelihood of states in $S$.
\end{Postulate} 
The constraints we will discuss later in the context of MaxEnt will typically be of this type:
constraints that describe empirical means of simple functions like polynomials of low order
have low complexity (provided that the constants involved have short descriptions), and 
 restrict the combinations of outcomes by huge factors. For the same reasons, typical constraints in thermodynamics also result from {\it mechanisms}: observing that all particles of a gas are located within a certain volume $V$
can only be explained by a mechanism (e.g. a wall) that confines them to $V$, rather than being just a coincidence.\footnote{In general, constraints on {\it macroscopic} variables have negligible description length compared to the typical
complexity of the microscopic state of a many-particle system, as also argued by \citet{ZurekKol}.}


\paragraph{Is the constraint $S$ tight?} 
Together with the bias for an ontic interpretation of constraints, we are now getting slightly closer to deriving
Postulate \ref{post:sep}  from IM (beyond the few comments made right after stating it). 
We now assume, for simplicity, that the constraint $(x,y)\in S$ is due to a mechanism that forces all pairs to lie in $S$
(although Postulate \ref{post:mech} is weaker in the sense that it only assumes a mechanism that increases the likelihood of $S$). 
The question we are facing is wether $S$ is {\it tight} in the sense that {\it all} pairs in $S$ will occur after sufficiently many repetitions of the same experiment. We then need to assume that $S$ originates from
separate constraints for $X$ and $F$ because otherwise we would need a mechanism that controls $X$ and $F$ {\it jointly}
by varying them in a way that enforces $(x,f(x)) \in S$, in contradiction to the independence of mechanisms, as 
sketched after Postulate \ref{post:sep}. 

For the case where $S$ is not tight and the mechanism generates only $(x,y)$ pairs in $S' \subset S$ (but we don't know $S'$) 
we still choose the update according to Postulate \ref{post:sep}
because this is the only possible choice for constraints on $X$ and $F$ that doesn't commit beyond the information we have. 

We summarize that assuming that a constraint $S$ arises from independent constraints for 
$\cX$ and $\cF$ is our inductive bias, which can be justified under appropriate conditions.

\section{From MaxEnt to Causal MaxEnt \label{sec:maxent}} 

\subsection{Wallis' argument for MaxEnt \label{subsec:wallis}}
Inferring underdetermined probability distributions by maximizing entropy subject to the available information is
a well-established  principle in machine learning and statistics, see e.g. \cite{Frogner2019,Levy1994,Myung1996}. The usual formal setting reads:

\paragraph{Accounting for linear constraints} Let us, for simplicity, assume that $X$ is a variable that attains values in
some finite set $\cX$. Assume the only  information available on $P_X$ is given by the expectations
\begin{equation}\label{eq:constr}
\sum p(x) f_j(X) = c_j, \quad \hbox{ with } c_j \in \R,
\end{equation}    
where $f_j$ are measurable functions. According to MaxEnt we would then choose the unique
distribution maximizing the Shannon entropy\footnote{For continuous variables, one typically replaces Shannon entropy with differential Shannon entropy \cite{cover} $H(X):= -\int p(x) \log p(x) dx$. Since the latter is not invariant 
with respect to re-parametrization, one should then rather consider minimization of relative entropy to a given prior distribution.}
\[
H(X):= - \sum_x p(x) \log p(x).
\]
subject to the constraints \eqref{eq:constr}, which yields
\begin{equation}\label{eq:maxentdistr}
p(x) = e^{- \lambda_j f_j(x) -\mu},
\end{equation}
with appropriate Lagrange multipliers $\lambda_j,\mu$.

While distributions that result from MaxEnt often appear intuitively 'natural', or 'smooth'\footnote{Since any distribution maximizes the entropy subject to appropriate constraints (just choose $f(x) := \log q(x)$ with appropriate constant $c$), this is certainly a result of the {\it type of constraints} that typically occur in applications, e.g., if only first and second moments of a distribution are known}, there is an ongoing debate about how to justify \eqref{eq:maxentdistr} as a {\it rational} guess  \cite{Jaynes1957,Palmieri2013,Uffink1996}.

\citet{Shore1978} stated Postulates that  'consistent' rules for updating a distribution after new information comes in should satisfy, \cite{Uffink1996} criticized the approach as suffering from hidden implicit assumptions that go beyond what \cite{Uffink1996} would call `consistency' requirements. 
We will therefore prefer the so-called Wallis' derivation (see \cite{Jaynes2003}, Section 11.4), which we briefly sketch: 
Consider an experiment with $n$ draws from the finite probability space
 $\cX=\{x^1,\dots,x^k\}$, and  $n_1,\dots,n_k$ with $\sum_j n_j =n$ denotes the number of occurrences of $x^j$. 
 By elementary  combinatorics, the  number of combinations for these frequencies reads
 \begin{equation}\label{eq:N} 
\# (n_1,\dots,n_k) = \frac{n!}{n_1 ! n_2 ! \cdots n_k!}. 
\end{equation} 
Using Stirling's approximation one can easily show  that
\begin{eqnarray}\label{eq:logN}
\frac{1}{n} \log \# (n_1,\dots,n_k) &=&- \sum_j \frac{n_j}{n} \log \frac{n_j}{n} \\
&+& O\left(\log n/n\right). \nonumber
\end{eqnarray}  
Hence, the number of realizations can be estimated via the entropy of the relative frequencies.
Thus, the MaxEnt distribution is the distribution for which the corresponding relative frequencies maximize the
number of realizations in the limit of $n\to \infty$. 

Further, one can show that for large enough $n$, the overwhelming majority of $n$-tuples satisfying the constraints
show empirical distributions that are close to the MaxEnt distribution. Hence, a prior on $\cX^n$ that assigns equal probability to each $n$-tuple, results, after accounting for the constraints, in a posterior that is essentially supported by empirical distributions close to the unique MaxEnt distribution. In this sense, MaxEnt can also be seen as an implication of PIR (when applied to empirical distributions), although MaxEnt is more general from the formal point of view.

\subsection{Causal MaxEnt from Causal PIR} 

We start with motivating Causal MaxEnt in the same way as it is done by \citet{SunLauderdale}. Assume we are given a 
continuous variable $X$ as cause and a binary variable $Y$ as effect. Let the only information about the joint distribution
$P_{X,Y}$ be given by the first and second moments $\Exp[X]$, $\Exp[X^2]$, $\Exp[XY]$, $\Exp[Y]$, $\Exp[Y^2]$. 
One can easily verify that the MaxEnt distribution is a bivariate mixture of Gaussians, where the cases $Y=0,1$ correspond 
to the two mixture components.  \citet{SunLauderdale} argue that this distribution would be a plausible distribution if $Y$ was the cause and $X$ the effect, while it is not plausible that the {\it cause} becomes bimodal just because it has an influence on a binary variable. If one, instead, first maximizes $H(X)$ subject to the constraints  $\Exp[X]$, $\Exp[X^2]$ and then $H(Y|X)$ 
subject to the remaining constraints $\Exp[XY]$, $\Exp[Y]$, $\Exp[Y^2]$, the marginal distribution $P_X$ becomes a single Gaussian and
$P_{Y=1|X}$ a sigmoid function where the probability for $Y=1$ smoothly increases or decreases with $X$, which 
 \citet{SunLauderdale} consider plausible for the causal direction $X\to Y$.
Formally, they have postulated the following principle:
\begin{Definition}[Causal MaxEnt] 
Given some linear constraints for $P_{X,Y}$ for the cause effect pair $(X,Y)$. 
Infer the bivariate distribution by first maximizing $H(X)$ subject to all constraints for $P_X$ (entailed by 
the joint constraints). 
Then, maximize $H(Y|X)$ subject to the joint constraints. 
\end{Definition} 
\citet{SecondOrder} show that usual MaxEnt violates the algorithmic independence of
$P_X$ and $P_{Y|X}$.  

The proof is based on the observation that MaxEnt can result in a joint distribution whose marginal
$P_X$ cannot be defined by a separate constraint with simple description. 
Instead, its simplest description may be `the marginal distribution
resulting from MaxEnt for the joint constraint'. For the example above with binary $X$ and real-valued $Y$ with second order constraints, $P_X$ is a mixture of two Gaussians, and thus already contains the full information about the joint distribution $P_{X,Y}$. 
Despite describing this problem of MaxEnt, \citet{SecondOrder} do not show that
Causal MaxEnt is the right replacement of MaxEnt.

 We now show that this sequential probability update is a result of Causal PIR when applied to empirical distributions.\footnote{For readers with interest in physics we note that the independent
uniform distributions on $\cX^n$ and $\cF^n$ our arguments in Subsection \ref{subsec:priorf} relied on 
can be seen as a result of independent mixing processes, the first one mixes the state of the cause and the second one the conditional state of the effect. Accordingly, \citet{MitArmen} have described a physical toy model for a cause-effect relation where this sequential entropy maximization follows from mixing processes that first affect the cause and then the interaction that generates the effect from the cause. }  
Assume we are given $\ell$ constraints of the form
\begin{equation}\label{eq:bivconstraint} 
\Exp [f_j(X,Y)] = c_j \hbox{ for } j=1,\dots,\ell.
\end{equation} 

Let us now interpret \eqref{eq:bivconstraint} as constraints for the empirical distribution after $n$ draws. 
For each pair
$(\bx,\by)$  of $n$-tuples $\bx:=(x_1,\dots,x_n)$ and $\by:=(y_1,\dots,y_n)$ we denote by $\Exp_{(\bx,\by)}$ the expectation induced by the corresponding empirical distribution of $(X,Y)$. 
Finally, we define
\begin{eqnarray} 
S &:=& \{(\bx,\by) \in \cX^n\times \cY^n \hbox{ with }  \nonumber \\
  && |\Exp_{(\bx,\by)} [f_j(X,Y)]  - c_j| \leq \epsilon\} ,
\end{eqnarray} 
with some arbitrarily small $\epsilon > 0$, which defines a   relaxation of \eqref{eq:bivconstraint} to ensure feasibility for sufficiently large $n$. 

Following our separation of constraints in Postulate \ref{post:sep},  we now define $S_{\bX}$ as the set of $n$-tuples $\bx$ for which there exists an $n$-tuple $\by$ such that $(\bx,\by)\in S$. Again, Causal PIR tells us to put a uniform prior
on   $S_{\bX}$. Following Subsection \ref{subsec:wallis},  the overwhelming majority of $n$-tuples $\bx$ in  $S_{\bX}$ are close to the distribution $P_X$ that
maximizes $H(X)$ subject to \eqref{eq:bivconstraint} being feasible for $Y$. 
For any $\bx\in S_{\bX}$ let $S_\bx$ denote the set of $n$-tuples $\by$ such that $(\bx,\by)\in S$. 
According to causal PIR, we put a uniform prior on $S_\bx$. We will again use \eqref{eq:logN}
to derive the conditional empirical distribution that is induced by the majority of the $\by\in S_\bx$.  

To this end, for any $\bx\in  S_{\bX}$ let 
$n_1^\bx,\dots,n_k^\bx$ denote the number of occurrences of the $k$ different elements of $\cX$. Further, for any $\by\in S_\bx$, let $n^i_j$ be the number of occurrences of the element $(i,j)$ in $\cX\times \cY$. For any collection $(n^i_{j})$ and any fixed $\bx$, the number of different $\by$ is given by 
\begin{equation}
\#(n_1^1,\dots,n^k_l) = \prod_{i=1}^k \frac{n^\bx_i!}{n^i_1 ! n^i_2 ! \cdots n^i_k!},  
\end{equation} 
since we need to apply \eqref{eq:N} for each element of $\cX$ and sample size $n_1^\bx$. 
Using the same arguments as for the derivation of \eqref{eq:logN} we obtain
\begin{eqnarray}
&& \frac{1}{n} \log \# (n^1,\dots,n^k_l) \nonumber \\ 
&=& -  \sum_i \frac{n_i^\bx}{n}  \sum_{j}  \frac{n^i_j}{n^\bx_i}  \log \frac{n^i_j}{n^\bx_i}.  \label{eq:condemp}\\
&+&  O\left(\log n /n \right). \nonumber
\end{eqnarray}  
Recalling that the conditional entropy of $Y$ given $X$ for any probability mass function $p(x,y)$ reads \citep{cover} 
\[
H(Y|X) = -\sum_{x,y} p(x) p(y|x) \log p(y|x),  
\]
we observe that \eqref{eq:condemp} is the conditional entropy of the empirical distribution. 
Accordingly, we conclude that, for any fixed $\bx$, the overwhelming majority of $n$-tuples $\by$ in $S_\bx$ are those whose empirical distributions are close to the distribution maximizing conditional entropy subject to \eqref{eq:bivconstraint}. 

The above arguments show that  first putting a uniform prior on $S_{\bX}$ and then, for fixed $\bx$, a uniform prior on
$S_\bx$ yields a joint distribution on $\cX^n \times \cY^n$ that is strongly concentrated on the set of $(\bx,\by)$-pairs whose empirical distribution is given by Causal MaxEnt. In contrast, classical MaxEnt would provide the most likely empirical distribution only if we put uniform prior on $S$, that is, if we use standard PIR.

\section{Generalization of Causal MaxEnt to arbitrary DAGs\label{sec:N}} 

Given a causally sufficient set of $N$ variables $X_1,\dots,X_N$, causally linked by the directed acyclic graph (DAG) $G$,
the causal Markov condition \citep{Spirtes1993,Pearl:00} implies that the joint distribution factorizes according to
\begin{equation}\label{eq:markov} 
P_{X_1,\dots,X_N} = \prod_{j=1}^N P_{X_j|PA_j}, 
\end{equation} 
where $P_{X_j|PA_j}$ denotes the conditional distribution of $X_j$, given its parents in $G$. 
If we are given multivariate constraints of the form
\begin{equation}\label{eq:multicon}
\Exp[ f_j(X_1,\dots,X_N)] = c_j,
\end{equation}  
 the arguments from Section \ref{sec:maxent} suggest to obtain the conditionals $P_{X_j|PA_j}$ by sequentially maximizing conditional entropy $H(X_j |  PA_j)$
 according to an ordering that is consistent with $G$, a procedure already proposed by \citet{SunLauderdale}.
 Since we construct the joint distribution as the product of the conditionals  $P_{X_j|PA_j}$, it is Markov relative to $G$ {\it by construction}.
 This seems to overcome a problem with classical MaxEnt:
 maximizing the joint entropy subject to \eqref{eq:multicon} does not necessarily result in an Markovian distribution, while maximizing entropy
 subject to \eqref{eq:multicon} and \eqref{eq:markov} is not a {\it convex} optimization problem and thus need not have a unique solution (as shown below for a toy example).  

Before describing problems with Causal MaxEnt for general DAGs, let us first consider an example where it makes sense.
Let $X_j$ be binary variables connected by the causal structure
\begin{equation}
X_1 \to X_2 \to \cdots \to X_N. 
\end{equation}  
Assume now we are given a constraint saying '$X_j=0$ implies $X_{j+1}=0$' for $j=1,\dots,N-1$. 
Intuitively, this corresponds to a mechanism that appends $0$ or $1$ to any binary word ending with $1$, but it appends only $0$ to words ending with $0$.
In other words, it rules out any binary word $(x_1,\dots,x_N)$  
containing the substring $01$. 
Classical MaxEnt would thus result in the uniform distribution over the $N+1$ binary words $0 \dots 0$, $10 \dots 0$, $110\cdots 0$,  
$\dots$, $11 \dots 1$. Causal MaxEnt yields $X_1=1$ with probability $1/2$, and all other $X_j$ attain $1$ with probability $1/2$ if their predecessor is $1$. Thus, the binary words occur with probability $1/2^1, 1/2^2,\dots, 1/2^N,1/2^N$, a distribution with much lower entropy. In this sense, Causal MaxEnt is more conclusive since it results in smaller uncertainty about the resulting joint distribution after levering the causal information.

 However, sequential entropy maximization raises the following two problems (ignored by  \citet{SunLauderdale}) in case
 the DAG is not complete:\footnote{A DAG is called complete if adding further arrows would result in directed cycles.}  
 First, the ordering of nodes is not necessarily unique. Second, sequentially maximizing entropy 
 may render the constraints \eqref{eq:multicon} infeasible, as shown by the following toy example with a DAG $G$ with two variables and no edge.
Consider the
  binary variables $X_1,X_2$
  with values $\pm 1$. The Markov condition implies the factorization 
  \begin{equation}\label{eq:2mk} 
  P_{X_1,X_2}=P_{X_1} P_{X_2}.
  \end{equation}  
  Assume we are given the constraint
 \begin{equation}\label{eq:binary}
\Exp[ X_1 X_2 ] = 1.
 \end{equation} 
 To implement Causal MaxEnt, let us choose the ordering $X_1,X_2$.
We observe that \eqref{eq:binary} entails no restriction for the marginal distributions of $X_1$, and thus maximizing $H(X_1)$ yields $P(X_1=1)=1/2$.
Then Causal MaxEnt advices us to maximize the entropy of $X_2$, given its parents in $G$ (which is the empty set), subject to \eqref{eq:binary}.
However, there is no $P_{X_2}$ such that $P_{X_1}P_{X_2}$ satisfies \eqref{eq:binary}, after we have already maximized the entropy of $X_1$.   
To satisfy the constraint, we need $X_2$ {\it depending} on $X_1$, which violates the Markov condition. 
The only joint distributions satisfying 
constraint \eqref{eq:binary} and  Markov condition \eqref{eq:2mk} are point measures on $(1,1)$ or $(-1,-1)$, respectively. These are the 
two solutions of the non-convex problem of maximizing entropy subject to \eqref{eq:binary} {\it and} \eqref{eq:2mk}. 
By deciding for one of the solutions we  would commit beyond the known constraint \eqref{eq:binary}. 
If \eqref{eq:binary} results from independent mechanisms for $X_1$ and $X_2$, it could be that there are either two independent mechanisms 
generating only the value $1$ for both variables, or independent mechanisms generating only $-1$ for both ones, we just do not know  which scenario is the true one.
In other words,  \eqref{eq:binary} represents our {\it knowledge} on the mechanisms, while the mechanisms themselves respect tighter constraints, namely 
$(X_1,X_2)=(1,1)$ or $(X_1,X_2)=(-1,-1)$, depending on the scenario.   
Hence we have an example for the case where constraints are not 'tight' in the sense of our discussion in Subsection \ref{subsec:just}.  

More generally speaking, the example shows that decomposing constraints like \eqref{eq:multicon}
 into {\it independent} constraints for each of the mechanisms $P_{X_j|PA_j}$ may not be possible. The bivariate example suggests that the requirement to obtain a distribution that factorizes according to the DAG structure 
 prohibits using the constraints entirely, given that we must not commit to any information that is not entailed by the constraints (as we would do by choosing either $(X_1,X_2)= (1,1)$ or $(X_1,X_2)= (-1,-1)$).

\section{Deriving Information Geometric Causal Inference from Causal PIR \label{sec:igci}} 

\begin{figure}
\resizebox{3.7cm}{!}{%
\begin{tikzpicture}
\draw[thick,->] (0,0) -- (4.5,0) node[anchor=north west] {X};
\draw[thick,->] (0,0) -- (0,4.5) node[anchor=south east] {Y};
\draw (4 cm,3pt) -- (4 cm,-3pt) node[anchor=north] {$1$};
\draw (3pt,4cm) -- (-3pt,4cm) node[anchor=east] {$1$};
\draw (0,0) .. controls (0,4) and (4,0) .. (4,4);
\node (f) at (4,4.3)  {$f$}; 
\draw[red,thick,fill=red] (2,2) circle [radius=0.1cm] ;
\draw[red,thick,fill=red] (1,1.89) circle [radius=0.1cm] ;
\draw[red,thick,fill=red] (3.45,2.34) circle [radius=0.1cm] ;
\end{tikzpicture}
}
\hfill
\resizebox{3.7cm}{!}{%
\begin{tikzpicture}
\draw[thick,->] (0,0) -- (4.5,0) node[anchor=north west] {X};
\draw[thick,->] (0,0) -- (0,4.5) node[anchor=south east] {Y};
\draw (4 cm,3pt) -- (4 cm,-3pt) node[anchor=north] {$1$};
\draw (3pt,4cm) -- (-3pt,4cm) node[anchor=east] {$1$};
\draw[line width=0.3cm] (0,0) .. controls (0,4) and (4,0) .. (4,4);
\node (f) at (4,4.3)  {$f$}; 
\draw[white,thick,fill=white] (2,2) circle [radius=0.1cm] ;
\draw[white,thick,fill=white] (1,1.89) circle [radius=0.1cm] ;
\draw[white,thick,fill=white] (3.45,2.34) circle [radius=0.1cm] ;
\foreach \x in {0,1,2,3,4,5,6,7,8,9,10,11,12,13,14,15,16,17,18,19,20,21,22,23,24,25,26,27,28}
    \foreach \y in {0,1,2,3,4,5,6,7,8,9,10,11,12,13,14,15,16,17,18,19,20,21,22,23,24,25,26,27,28}
       \draw  (\x/7, \y/7) circle [radius=0.01cm];     
\end{tikzpicture}
}
\caption{\label{fig:igci} Left: IGCI for a bijective function $f:[0,1]\to [0,1]$. Under certain genericity assumptions, $x$-values tend to lie in regions with small slope $f'(x)$. Right: Drawing the function $f$ with a fat pen, it induces a relation of possible $(x,y)$-pairs on the grid (obtained by discretizing $X$ and $Y$).} 
\end{figure}
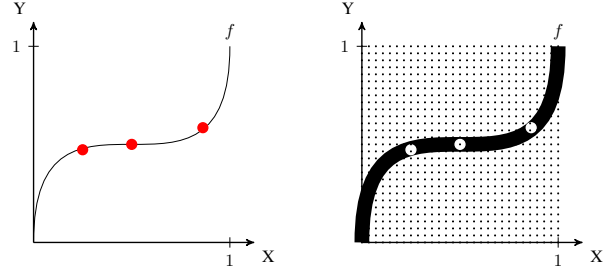

Information Geometric Causal Inference (IGCI)  \citep{deterministic,Info-Geometry} is a method for causal discovery that infers whether two $X$ causes $Y$ for $Y$ causes $X$ from the bivariate distribution $P_{X,Y}$ for the case of an invertible {\it deterministic} causal relation,  i.e., $Y=f(X)$ and $X=f^{-1}(Y)$. Although IGCI is more general, we sketch the idea for variables with range $[0,1]$
and strictly monotonously increasing $f$, as shown in Figure \ref{fig:igci}, left.   
The intuitive idea is that, for the causal relation $X\to Y$, 'generic choices' of $P_X$ (independently chosen of $f$)\footnote{formalized by the condition $\int_0^1 \log f'(x) p(x) dx \approx \int_0^1 \log f'(x) dx $.} result in distributions $P_Y$ that tend to have higher density in regions where the derivative $(f^{-1})'(y)$ is large.   
To exploit this asymmetry for inferring the direction, one infers $X\to Y$ iff points accumulate in regions of small $f'$ rather than small  $f^{-1'}$. 
Formally, IGCI amounts to inferring the direction $X\to Y$ iff\footnote{note the symmetry $\sum_{j=1}^n \log f'(x_j)= - \sum_{j=1}^n \log f^{-1'}(y_j)$.}
\[
\sum_{j=1}^n \log f'(x_j) < 0. 
\]
IGCI can be obtained as the deterministic and continuous limit of Causal PIR in the following sense.
Note that our derivation is close in spirit to the justification of IGCI provided by
\citet{JustiIGCI}, which relies on counting arguments in the space of discrete functions. However, we want to directly derive it from Causal PIR.

Assume we draw the function $f$ with a fat pen, as shown in
Figure \ref{fig:igci}, right.
Define, after discretizing $X$ and $Y$ to get a grid with $\ell \times \ell$ points,
define  $R \subset \cX \times \cY$ as the points $(x,y)$ lying on the fat stripe.  
For each $x$, let $N_X(x)$ denote the number of possible $y$-values for which $(x,y)\in R$. Define $N_Y(y)$ similarly. 
For each observed point $(x_j,y_j)$ we 
 have
\begin{equation}\label{eq:slope}
f'(x_j) \approx \frac{N_X(x_j)}{N_Y(y_j)}.   
\end{equation}
Then,  $\prod_{j=1}^n N_X(x_j) $
 is the number of possible $n$-tuples $\by$ for the observed
$\bx$. Likewise, $\prod_{j=1}^n N_Y(y_j) $,
 is the number of possible $n$-tuples $\bx$ for the observed $\by$.

On checks easily that inferring causal direction via Causal PIR based cause-effect inference in Definition \ref{def:MLCausalPIR} 
thus amounts to comparing 
$\sum_{j=1}^n \log N_Y(y_j)$ to $\quad \sum_{j=1}^n \log N_X(x_j) $, which, after using \eqref{eq:slope} amounts to
checking the sign of  $\sum_{j=1}^n \log f'(x_j)$.   

We have thus shown that another non-trivial causal inference method (part from Causal MaxEnt) also follows from applying Causal PIR to the $n$-fold cartesian product of the underlying probability space.

\section{Conclusions}

Using a simple mechanical device as toy example, we have argued that our common sense replaces PIR with Causal PIR
for bivariate distributions of cause and effect whenever we account for knowledge on the mechanism connecting cause and effect.  We have further justified Causal PIR and Causal MaxEnt by assuming
that constraints on joint distributions arise from two separate constraints: constraints  on the cause and constraints on the cause-effect relation (in the sense of possible functions). 
Earlier work has solved paradoxes with usual MaxEnt by updating priors over functions, too.
We have argued, however, that knowledge on the bivariate distribution is not a priori divided into information on the cause and information on the functional relation between cause and effect. We have therefore proposed a way to divide it into these two components in a way that entails minimal commitment for both of them.

\paragraph{Acknowledgements} Many thanks to Armen Allahverdyan and Sergio Hernan Garrido Mejia for helpful comments on relevant literature and Elke Kirschbaum for remarks on an earlier version.


\begin{thebibliography}{39}
\providecommand{\natexlab}[1]{#1}
\providecommand{\url}[1]{\texttt{#1}}
\expandafter\ifx\csname urlstyle\endcsname\relax
  \providecommand{\doi}[1]{doi: #1}\else
  \providecommand{\doi}{doi: \begingroup \urlstyle{rm}\Url}\fi

\bibitem[Allahverdyan and Janzing(2008)]{MitArmen}
A.~Allahverdyan and D.~Janzing.
\newblock Relating the thermodynamic arrow of time to the causal arrow.
\newblock \emph{J.~Stat.~ Mech.}, 4:\penalty0 P04001, 2008.

\bibitem[Bengio et~al.(2019)Bengio, Deleu, Rahaman, Ke, Lachapelle, Bilaniuk,
  Goyal, and Pal]{Bengio2019}
Y.~Bengio, T.~Deleu, N.~Rahaman, R.~Ke, S.~Lachapelle, O.~Bilaniuk, A.~Goyal,
  and C.~Pal.
\newblock A meta-transfer objective for learning to disentangle causal
  mechanisms, 2019.
\newblock URL \url{arXiv:1901.10912}.

\bibitem[Blöbaum et~al.(2017)Blöbaum, Washio, and Shimizu]{Bloebaum17}
P.~Blöbaum, T.~Washio, and S.~Shimizu.
\newblock Error asymmetry in causal and anticausal regression.
\newblock \emph{Behaviormetrika}, pages 1--22, 2017.

\bibitem[Cover and Thomas(1991)]{cover}
T.~Cover and J.~Thomas.
\newblock \emph{Elements of Information Theory}.
\newblock Wileys Series in Telecommunications, New York, 1991.

\bibitem[Daniusis et~al.(2010)Daniusis, Janzing, Mooij, Zscheischler, Steudel,
  Zhang, and Sch{\"o}lkopf]{deterministic}
P.~Daniusis, D.~Janzing, J.~M. Mooij, J.~Zscheischler, B.~Steudel, K.~Zhang,
  and B.~Sch{\"o}lkopf.
\newblock Inferring deterministic causal relations.
\newblock In \emph{Proceedings of the 26th Annual Conference on {U}ncertainty
  in {A}rtificial {I}ntelligence ({UAI})}, pages 143--150. AUAI Press, 2010.

\bibitem[Frogner and Poggio(2019)]{Frogner2019}
C.~Frogner and T.~Poggio.
\newblock Fast and flexible inference of joint distributions from their
  marginals.
\newblock In \emph{International Conference on Machine Learning}, 2019.

\bibitem[Guyon et~al.(2019)Guyon, Statnikov, and Bakır-Batu]{Guyon2019}
I.~Guyon, A.~Statnikov, and B.~Bakır-Batu.
\newblock \emph{Cause Effect Pairs in Machine Learning}.
\newblock The Springer Series on Challenges in Machine Learning. Springer,
  Berlin \& Heidelberg, 01 2019.

\bibitem[Hoyer et~al.(2009)Hoyer, Janzing, Mooij, Peters, and
  Sch\"olkopf]{Hoyer}
P.~Hoyer, D.~Janzing, J.~Mooij, J.~Peters, and B.~Sch\"olkopf.
\newblock Nonlinear causal discovery with additive noise models.
\newblock In D.~Koller, D.~Schuurmans, Y.~Bengio, and L.~Bottou, editors,
  \emph{Proceedings of the conference Neural Information Processing Systems
  (NIPS) 2008}, Vancouver, Canada, 2009. MIT Press.

\bibitem[Hunter(1986)]{Hunter1986}
D.~Hunter.
\newblock Uncertain reasoning using maximum entropy inference.
\newblock In \emph{Uncertainty in Artificial Intelligence}, volume~4 of
  \emph{Machine Intelligence and Pattern Recognition}, pages 203 -- 209.
  North-Holland, 1986.

\bibitem[Hunter(1989)]{Hunter1989}
D.~Hunter.
\newblock Causality and maximum entropy updating.
\newblock \emph{International Journal of Approximate Reasoning}, 3\penalty0
  (1):\penalty0 87 -- 114, 1989.

\bibitem[Janzing(2019)]{misconceptions}
D.~Janzing.
\newblock The cause-effect problem: Motivation, ideas, and popular
  misconceptions.
\newblock In I.~Guyon, R.~Statnikov, and B.~Bakir~Batu, editors, \emph{Cause
  Effect Pairs in Machine Learning}, pages 3--26. Springer, 2019.

\bibitem[Janzing and Sch\"olkopf(2010)]{Algorithmic}
D.~Janzing and B.~Sch\"olkopf.
\newblock {Causal inference using the algorithmic Markov condition}.
\newblock \emph{IEEE Transactions on Information Theory}, 56\penalty0
  (10):\penalty0 5168--5194, 2010.

\bibitem[Janzing et~al.(2009)Janzing, Sun, and Sch\"olkopf]{SecondOrder}
D.~Janzing, X.~Sun, and B.~Sch\"olkopf.
\newblock Distinguishing cause and effect via second order exponential models.
\newblock \emph{\em{\url{http://arxiv.org/abs/0910.5561}}}, 2009.

\bibitem[Janzing et~al.(2012)Janzing, Mooij, Zhang, Lemeire, Zscheischler,
  Daniu\v{s}is, Steudel, and Sch\"olkopf]{Info-Geometry}
D.~Janzing, J.~Mooij, K.~Zhang, J.~Lemeire, J.~Zscheischler, P.~Daniu\v{s}is,
  B.~Steudel, and B.~Sch\"olkopf.
\newblock Information-geometric approach to inferring causal directions.
\newblock \emph{Artificial Intelligence}, 182--183:\penalty0 1--31, 2012.

\bibitem[Janzing et~al.(2015)Janzing, Steudel, Shajarisales, and
  Sch\"olkopf]{JustiIGCI}
D.~Janzing, B.~Steudel, N.~Shajarisales, and B.~Sch\"olkopf.
\newblock {Justifying information-geometric causal inference}.
\newblock In V.~Vovk, P.~H., and A.~Gammerman, editors, \emph{Measures of
  Complexity}, Festschrift for Alexey Chervonencis, pages 253--265. Springer
  Verlag, Heidelberg, 2015.

\bibitem[Janzing et~al.(2016)Janzing, Chaves, and Sch\"olkopf]{AICarrowoftime}
D.~Janzing, R.~Chaves, and B.~Sch\"olkopf.
\newblock Algorithmic independence of initial condition and dynamical law in
  thermodynamics and causal inference.
\newblock \emph{New Journal of Physics}, 18\penalty0 (093052):\penalty0 1--13,
  2016.

\bibitem[Jaynes(2003)]{Jaynes2003}
E.~Jaynes.
\newblock \emph{Probability theory: the logic of science}.
\newblock Cambridge University Press, Cambridge, MA, 2003.

\bibitem[Jaynes(1957)]{Jaynes1957}
E.~T. Jaynes.
\newblock Information theory and statistical mechanics. ii.
\newblock \emph{Phys. Rev.}, 108:\penalty0 171--190, Oct 1957.

\bibitem[Kano and Shimizu(2003)]{Kano2003}
Y.~Kano and S.~Shimizu.
\newblock Causal inference using nonnormality.
\newblock In \emph{Proceedings of the International Symposium on Science of
  Modeling, the 30th Anniversary of the Information Criterion}, pages 261--270,
  Tokyo, Japan, 2003.

\bibitem[Kocaoglu et~al.(2017)Kocaoglu, Dimakis, Vishwanath, and
  Hassibi]{Kocaoglu2017}
M.~Kocaoglu, A.~G. Dimakis, S.~Vishwanath, and B.~Hassibi.
\newblock Entropic causal inference.
\newblock In \emph{Proceedings of the Thirty-First AAAI Conference on
  Artificial Intelligence}, AAAI'17, page 1156–1162. AAAI Press, 2017.

\bibitem[Lemeire and Janzing(2012)]{LemeireJ2012}
J.~Lemeire and D.~Janzing.
\newblock Replacing causal faithfulness with algorithmic independence of
  conditionals.
\newblock \emph{Minds and Machines}, 23\penalty0 (2):\penalty0 227--249, 7
  2012.

\bibitem[{Levy} and {Delic}(1994)]{Levy1994}
W.~{Levy} and H.~{Delic}.
\newblock Maximum entropy aggregation of individual opinions.
\newblock \emph{IEEE Transactions on Systems, Man, and Cybernetics},
  24\penalty0 (4):\penalty0 606--613, 1994.

\bibitem[Li and Vit\'{a}nyi(1997)]{Vitanyi08}
M.~Li and P.~Vit\'{a}nyi.
\newblock \emph{An Introduction to Kolmogorov Complexity and its Applications}.
\newblock Springer, New York, 1997.

\bibitem[Marx and Vreeken(2017)]{Marx2017}
A.~Marx and J.~Vreeken.
\newblock Telling cause from effect using mdl-based local and global
  regression.
\newblock In \emph{2017 {IEEE} International Conference on Data Mining, {ICDM}
  2017, New Orleans, LA, USA, November 18-21, 2017}, pages 307--316, 2017.

\bibitem[Mooij et~al.(2016)Mooij, Peters, Janzing, Zscheischler, and
  Sch\"olkopf]{Mooij2016}
J.~Mooij, J.~Peters, D.~Janzing, J.~Zscheischler, and B.~Sch\"olkopf.
\newblock Distinguishing cause from effect using observational data: methods
  and benchmarks.
\newblock \emph{Journal of Machine Learning Research}, 17\penalty0
  (32):\penalty0 1--102, 2016.

\bibitem[Myung et~al.(1996)Myung, Ramamoorti, and Bailey]{Myung1996}
I.~J. Myung, S.~Ramamoorti, and A.~D. Bailey.
\newblock Maximum entropy aggregation of expert predictions.
\newblock \emph{Management Science}, 42\penalty0 (10):\penalty0 1420--1436,
  1996.

\bibitem[Palmieri and Domenico(2013)]{Palmieri2013}
F.~Palmieri and C.~Domenico.
\newblock Objective priors from maximum entropy in data classification.
\newblock \emph{Information Fusion}, 14\penalty0 (2):\penalty0 186 -- 198,
  2013.

\bibitem[Pearl(2000)]{Pearl:00}
J.~Pearl.
\newblock \emph{Causality}.
\newblock Cambridge University Press, 2000.

\bibitem[Peters et~al.(2017)Peters, Janzing, and Sch\"olkopf]{causality_book}
J.~Peters, D.~Janzing, and B.~Sch\"olkopf.
\newblock \emph{Elements of Causal Inference -- Foundations and Learning
  Algorithms}.
\newblock MIT Press, 2017.

\bibitem[Sch\"olkopf et~al.(2012)Sch\"olkopf, Janzing, Peters, Sgouritsa,
  Zhang, and Mooij]{anticausal}
B.~Sch\"olkopf, D.~Janzing, J.~Peters, E.~Sgouritsa, K.~Zhang, and J.~Mooij.
\newblock On causal and anticausal learning.
\newblock In L.~J. and J.~Pineau, editors, \emph{Proceedings of the 29th
  International Conference on Machine Learning (ICML)}, pages 1255--1262. ACM,
  2012.

\bibitem[Shajarisales et~al.(2015)Shajarisales, Janzing, Sch\"olkopf, and
  Janzing]{sic_icml}
N.~Shajarisales, D.~Janzing, B.~Sch\"olkopf, and D.~Janzing.
\newblock Telling cause from effect in deterministic linear dynamical systems.
\newblock In \emph{Proceedings of the International Conference on Machine
  Learning}, Lille, 2015.
\newblock to appear.

\bibitem[Shore and Johnson(1978)]{Shore1978}
J.~Shore and R.~Johnson.
\newblock Axiomatic derivation of the principle of maximum entropy and the
  principle of minimum cross-entropy.
\newblock \emph{Naval Research Laboratory, Washington D.C.}, pages 1--59, 1978.

\bibitem[Spirtes et~al.(1993)Spirtes, Glymour, and Scheines]{Spirtes1993}
P.~Spirtes, C.~Glymour, and R.~Scheines.
\newblock \emph{Causation, Prediction, and Search}.
\newblock Springer-Verlag, New York, NY, 1993.

\bibitem[Sun et~al.(2006)Sun, Janzing, and Sch\"{o}lkopf]{SunLauderdale}
X.~Sun, D.~Janzing, and B.~Sch\"{o}lkopf.
\newblock {Causal inference by choosing graphs with most plausible Markov
  kernels}.
\newblock In \emph{Proceedings of the 9th International Symposium on Artificial
  Intelligence and Mathematics}, pages 1--11, Fort Lauderdale, FL, 2006.

\bibitem[Uffink(1995)]{Uffink1995}
J.~Uffink.
\newblock Can the maximum entropy principle be explained as a consistency
  requirement?
\newblock \emph{Studies in History and Philosophy of Science Part B},
  26\penalty0 (3):\penalty0 223 -- 261, 1995.

\bibitem[Uffink(1996)]{Uffink1996}
J.~Uffink.
\newblock The constraint rule of the maximum entropy principle.
\newblock \emph{Studies in History and Philosophy of Science Part B: Studies in
  History and Philosophy of Modern Physics}, 27\penalty0 (1):\penalty0 47 --
  79, 1996.

\bibitem[Zhang and Hyv\"arinen(2009)]{Zhang_UAI}
K.~Zhang and A.~Hyv\"arinen.
\newblock On the identifiability of the post-nonlinear causal model.
\newblock In \emph{Proceedings of the 25th Conference on Uncertainty in
  Artificial Intelligence}, Montreal, Canada, 2009.

\bibitem[Ziebart et~al.(2013)Ziebart, Bagnell, and Dey]{Ziebart2013}
B.~Ziebart, J.~Bagnell, and A.~Dey.
\newblock The principle of maximum causal entropy for estimating interacting
  processes.
\newblock \emph{IEEE Transactions on Information Theory}, 59\penalty0
  (4):\penalty0 1966--1980, February 2013.

\bibitem[Zurek(1989)]{ZurekKol}
W.~Zurek.
\newblock Algorithmic randomness and physical entropy.
\newblock \emph{Phys Rev A}, 40\penalty0 (8):\penalty0 4731--4751, 1989.

\end{thebibliography}

\appendix

\section{Relation to Maximum Causal Entropy} 

\citet{Ziebart2013} consider a scenario with two interacting time series $(X_t)_{t\in \Z},(Y_t)_{t\in \Z}$ where
the latter ('predicted process') is to be inferred from the former ('known process'). Further, it is assumed that there are constraints (e.g. moment restrictions) 
on the joint distribution of the bivariate process capturing the relation between the two time series. Maximal Causal Entropy describes a way how an agent optimally accounts for observations from the known process for predicting the other process: 
While {\it future} values $X_s$ for $s>t$ can also contain information about the current value $Y_t$, \citet{Ziebart2013} 
sequentially maximize the entropy  of $Y_t$ subject to constraints referring to observations of $X_s$ for $s\leq t$, rather than accounting also for constraints that involve the entire process $(X_t)_{t \in \Z}$. The obvious argument is that
observations from $X_s$ for $s>t$ are not available at $t$ (which we will criticize below). 

Let us first mention an important conceptual difference to Causal MaxEnt.  
Maximum Causal Entropy a priori restricts the set of joint distributions over which is maximized. 
This is because it imposes conditional independences since every $Y_t$ is independent of future values $X_s$ ($s>t$), given the past of $Y$ plus the past and present of $X$. This a priori restriction is not made in Causal MaxEnt.  

\begin{figure}
\centerline{
\begin{tikzpicture}[scale=0.8, shorten >=1pt, shorten <=1pt]
\small
  \node[observed] at (0,0) (a0)  {$Y_{t-3}$};
  \node[observed] at (2,0) (a1)  {$Y_{t-2}$};
  \node[observed] at (4,0) (a2)  {$Y_{t-1}$};
  \node[observed] at (6,0) (a3)  {$Y_{t}$};
  \node[observed] at (0,2) (c0)  {$X_{t-3}$};
  \node[observed] at (2,2) (c1)  {$X_{t-2}$};
  \node[observed] at (4,2) (c2)  {$X_{t-1}$};
  \node[observed] at (6,2) (c3)  {$X_{t}$};
 \draw[-arcsq] (a2) -- (a3); 
  \draw[-arcsq] (c3) -- (a3); 
\end{tikzpicture}
}
\caption{\label{fig:ts} Interaction between two time series in which $X_t$ controls the mechanism relating
$Y_{t-1}$ and $Y_t$, but all other observations are independent.} 
\end{figure}
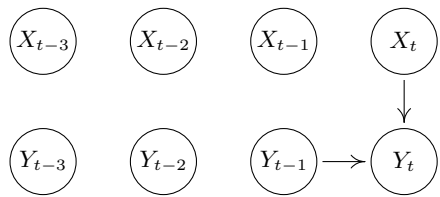

To explain our problems with the justification of Maximum Causal Entropy, we describe a scenario where Maximum Causal Entropy amounts to our Causal MaxEnt and Causal PIR, but with different justification. 
Consider the case where the interaction is as in Figure \ref{fig:ts}. 
For some fixed $t$, $Y_t$ is influenced by $Y_{t-1}$  and $X_t$. Assume that $Y_t$ can attain the values $1,2,3$ 
as our variables $X,Y$ in Section \ref{sec:PIR}. Further, let $X_t$ be binary and assume that its influence consists in switching between two different mechanisms for the relation between $Y_{t-1}$ and $Y_t$ : whenever $X_t=1$, 
$Y_{t-1}$ and $Y_t$ are related by the mechanical device in Section \ref{sec:PIR}. Whenever $X_t=0$, $Y_t$ is drawn independently of $Y_{t-1}$.   
To infer the joint distribution of $Y_{t-1},Y_t$, Maximum Causal Entropy  first chooses $P_{Y_{t-1}}$ to be uniform, since
the mechanism relating $Y_{t-1}$ and $Y_t$ is not known at that time. Then, after observing $X_t$, it constructs
$P_{Y_{t}|Y_{t-1}, X_t}$  as the uniform distribution over all $Y_t$ allowed by the mechanism determined by $X_t$. 
Here, the resulting joint distribution coincides with the one constructed via  Causal PIR and Causal MaxEnt for both cases
$X_t=0,1$. 

However, we do not believe that \citet{Ziebart2013} answer the question of {\it why} one should account for the known constraints in this sequential way. The fact that $X_t$ is {\it not known} when $Y_{t-1}$ is inferred, does not justify 
to entirely ignore the knowledge on the mechanisms. After all, we know that there are $13= 4 + 9$ possible combinations
for the triple $(y_{t-1},y_t,x_t)$. Assigning a uniform prior over them would result in a marginal distribution $P_{Y_{t-1}}$
that still slightly prefers the value $Y_{t-1}$ because it offers more options for $Y_t$ in case $X_t$ attains $1$ (and equally many options otherwise). In order words, from the point of view of optimally using all available information, we could, in any step, also account for constraints that refer to variables whose values are not known at that point in time and take their uncertainty in account.  
We believe that arguments similar to the ones in the present paper are required to complement the justification by \citet{Ziebart2013}.

\end{document}